\documentclass[10pt, conference, compsocconf]{IEEEtran}

\newcommand{\plainup}[1]{\textsuperscript{\normalfont#1}}  % 标准正体上标

\usepackage{graphicx} 
\usepackage{booktabs} % 专业表格排版
\usepackage{multirow} % 多行合并单元格
\usepackage[table]{xcolor} % 单元格颜色控制
\usepackage{xcolor} % 在导言区添加
\usepackage{makecell} % 支持\makecell命令
\ifCLASSINFOpdf
\else
\fi

\begin{document}
\title{The Effects of Communication Delay on Human Performance and Neurocognitive Responses in Mobile Robot Teleoperation}

\author{\IEEEauthorblockN{Zhaokun Chen\plainup{1},
Wenshuo Wang\plainup{1*},
Wenzhuo Liu\plainup{2}, 
Yichen Liu\plainup{2},
Junqiang Xi\plainup{1*}}
\IEEEauthorblockN{1. School of Mechanical Engineering, Beijing Institute of Technology, Beijing, China\\ 2. Beijing Institute of Technology, Zhuhai, China\\ Email: zk.chen@bit.edu.cn, ws.wang@bit.edu.cn, wzliu@bit.edu.cn, yc.liu@bit.edu.cn, xijunqiang@bit.edu.cn}
}

\maketitle

\begin{abstract}
Communication delays in mobile robot teleoperation adversely affect human-machine collaboration. Understanding delay effects on human operational performance and neurocognition is essential for resolving this issue. However, no previous research has explored this. To fill this gap, we conduct a human-in-the-loop experiment involving 10 participants, integrating electroencephalography (EEG) and robot behavior data under varying delays (0–500 ms in 100 ms increments) to systematically investigate these effects. Behavior analysis reveals significant performance degradation at 200–300 ms delays, affecting both task efficiency and accuracy. EEG analysis discovers features with significant delay dependence: frontal $\theta$/$\beta$-band and parietal $\alpha$-band power. We also identify a threshold window (100–200 ms) for early perception of delay in humans, during which these EEG features first exhibit significant differences. When delay exceeds 400 ms, all features plateau, indicating saturation of cognitive resource allocation at physiological limits. These findings provide the first evidence of perceptual and cognitive delay thresholds during teleoperation tasks in humans, offering critical neurocognitive insights for the design of delay compensation strategies.

\end{abstract}


\IEEEpeerreviewmaketitle

\section{Introduction}
Teleoperation is defined as the real-time remote control of robots by a human operator through communication networks \cite{lichiardopol2007survey}. This technology enables mobile robots to execute tasks in extreme environments inaccessible to human operators, with critical applications in disaster rescue \cite{aguilar2024development}, extraterrestrial exploration \cite{abubakar2024predictor}, and remote combat \cite{kot2018application}.
However, communication delays \cite{kamtam2024network}—caused by bandwidth constraints, excessive transmission distances, or signal processing requirements—can disrupt real-time synchronization between the operator and the robot, leading to degradation or even failure of the human-machine system performance \cite{kim2021impact}. The delay compensation algorithms presents a promising solution \cite{lu2019workload}, and an accurate understanding of how communication delays affect teleoperation is crucial for their development \cite{blackett2021effects}. 

Current studies focus on quantifying the effect of delay on mobile robot teleoperation by analyzing robot behavior features that reflect human performance (e.g., path deviation \cite{zheng2020evaluation}, collision rate \cite{lim2025communication}, and average speed \cite{neumeier2019teleoperation}). However, such features fail to reveal the effect of delay on human neurocognition responses (e.g., early perception or cognitive resource allocation strategies). Since neurocognitive processes directly drive behavioral responses \cite{ridderinkhof2014neurocognitive}, understanding these responses enables delay compensation algorithms to provide more appropriate compensation timing and intensity modulation parameters compared to only behavior features. To the best of our knowledge, no research in the field of mobile robot teleoperation has investigated these human neurocognitive responses.

To fill this research gap, our study establishes a simulation platform and designs a human-in-the-loop teleoperation experiment with varying delays. By employing statistical analysis methods, we systematically investigate the effects of communication delay on mobile robot teleoperation by integrating EEG and robot behavior data. The contributions of our work can be summarized as follows:

1) Analyzing the effect of delay on teleoperation task efficiency and accuracy using robot behavior features, and identifying the range of thresholds that lead to the first significant degradation of human performance.

2) Discovering EEG features with significant delay dependence, and determining the threshold for early perception of delay and the ceiling threshold for cognitive resource allocation in humans.

\begin{figure}[t]
    \centering
    \includegraphics[width=0.85\linewidth]{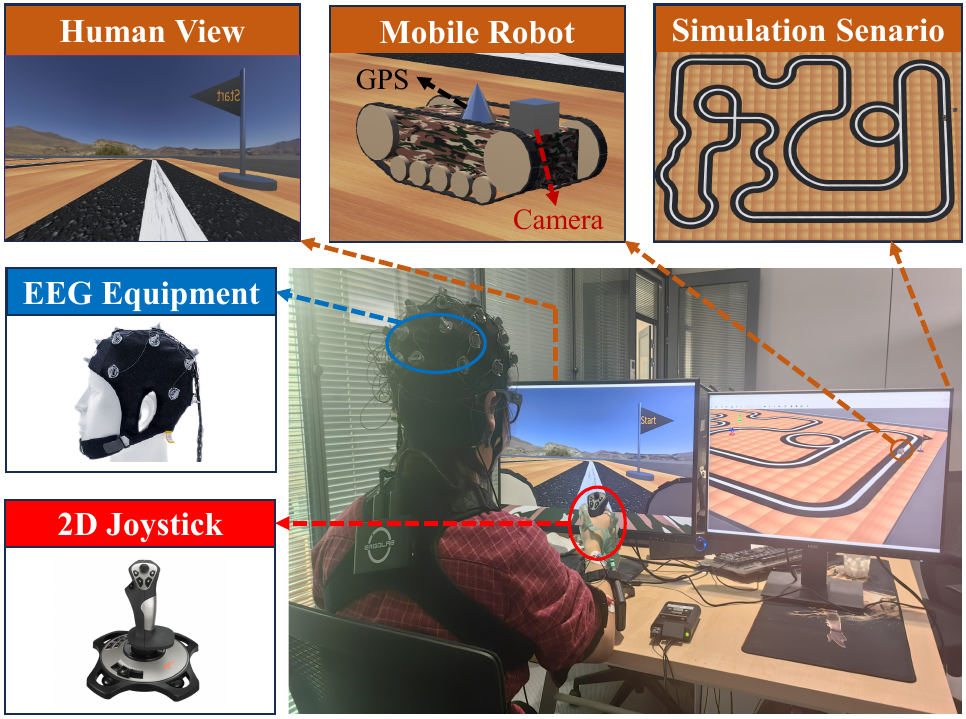}
    \caption{Illustration of the experimental setup for teleoperation.}
    \label{fig: 实验平台图}
\end{figure}

\section{Experiment design}

\subsection{Experimental Setup}
In our experiment, $10$ participants ($7$ males and $3$ females) volunteered for testing (mean age = $23.2$, standard deviation = $1.93$). We implemented the teleoperation simulation based on the experimental setup illustrated in Fig. \ref{fig: 实验平台图}. The mobile robot is controlled using a 2D joystick in the simulated environment, while EEG signals are recorded using a 32-channel cap from Kingfar (a Chinese technology company). 

The virtual environment was developed using Webots (an open-source robotic simulation platform), while the kinematic model of the mobile robot was designed in MATLAB/Simulink (R2020b, MathWorks). The control algorithm for robot manipulation through a 2D joystick was implemented based on the methodology described in \cite{lee2022operator}.

To investigate the effects of communication delay, six delay conditions (0 ms, 100 ms, 200 ms, 300 ms, 400 ms, and 500 ms) were introduced into the control signal pathway between the joystick and the mobile robot. These delays were systematically applied using dedicated ``Delay'' modules in Simulink.

\begin{figure}[t]
    \centering
    \includegraphics[width=0.8\linewidth]{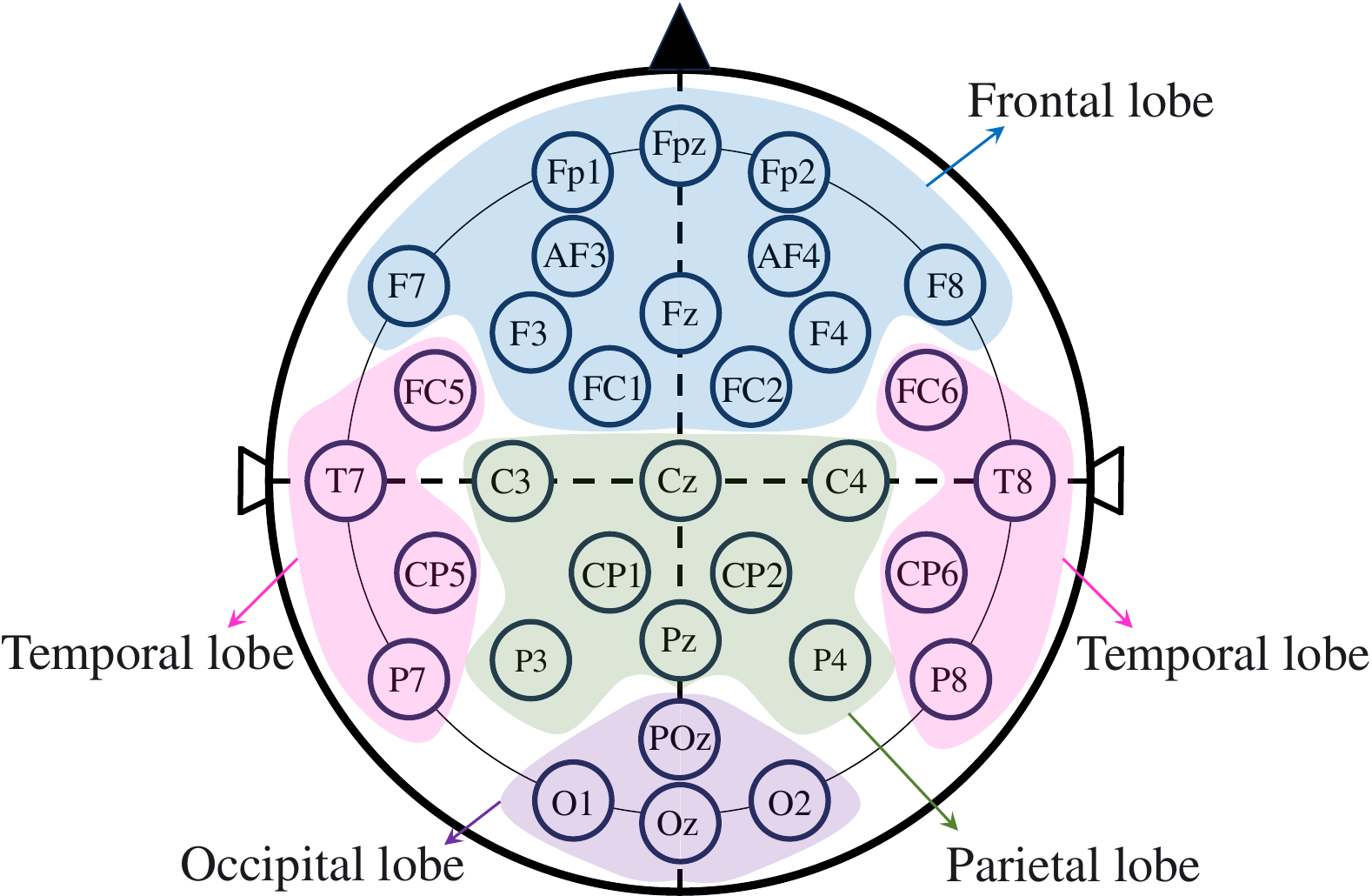}
    \caption{EEG channels and regions.}
    \label{fig: egg}
\end{figure}

\subsection{Experimental Procedure}

\subsubsection{Pre-Experiment Preparation}
Prior to the experiment, participants were briefed on the objectives and trained to operate the teleoperation system, ensuring uniform comprehension of the experimental requirements.

\subsubsection{Adaptation Phase}
Participants practiced under a 0 ms delay to adapt to the robot control, minimizing learning effects during main trials.

\subsubsection{Main Experiment}
Participants remotely controlled the mobile robot along the route of the simulation scenario in Fig. \ref{fig: 实验平台图} under six delay conditions (0 ms, 100 ms, 200 ms, 300 ms, 400 ms, and 500 ms), with the goal of finishing the task as quickly and accurately as possible.

\subsubsection{Data Collection}
Robot behavior (e.g., speed and trajectory coordinates) and EEG data were recorded synchronously ($10$ Hz) during each trial to analyze how varying magnitudes of communication delay affected mobile robot teleoperation.

\begin{figure}[t]
    \centering
    \includegraphics[width=0.85\linewidth]{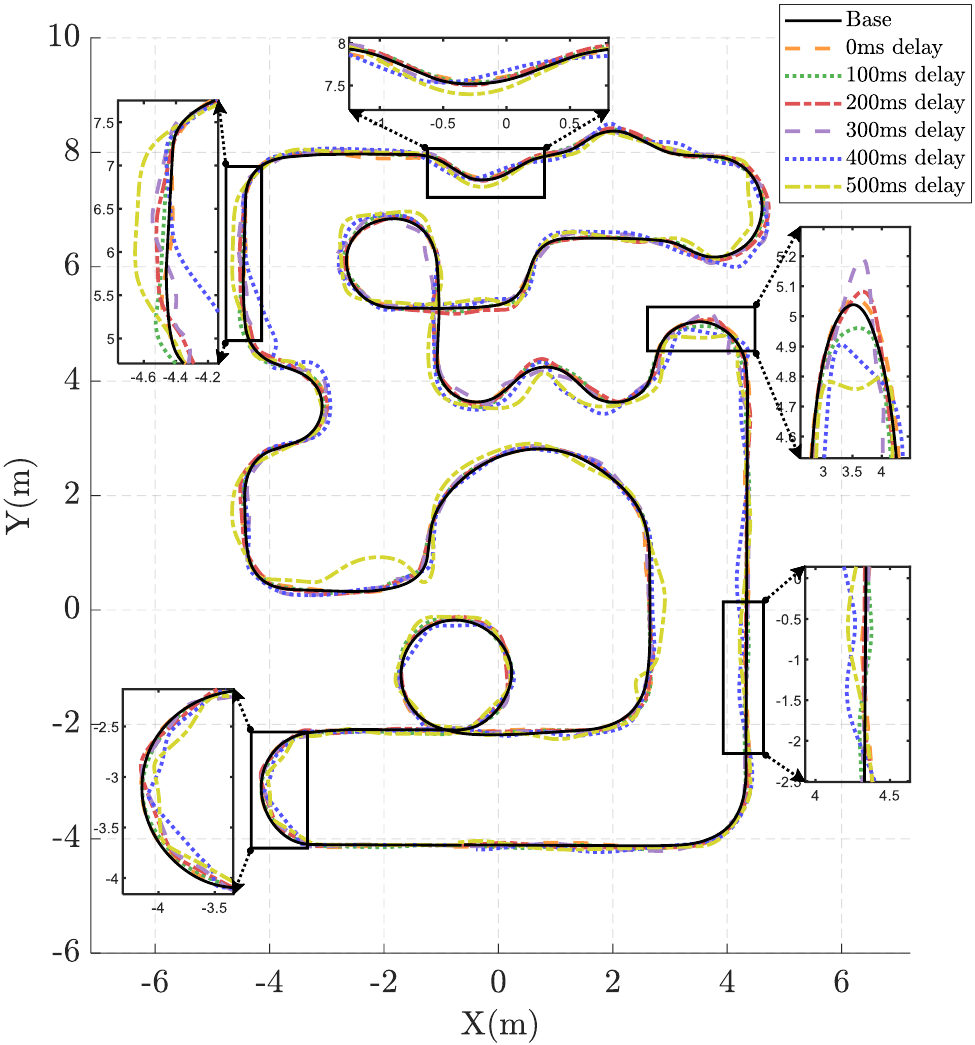}
    \caption{Trajectories of a representative participant under varying delay conditions compared to the baseline path.}
    \label{fig: trajectory}
\end{figure}

\subsection{Data Processing and Feature Extraction}
\subsubsection{Robot behavior data}
Based on the robot speed and its trajectory coordinates, we extracted average speed (efficiency metric) and Mean Lateral Position Deviation (MLPD) \cite{neumeier2019teleoperation} (accuracy metric) as robot behavior features. Fig. \ref{fig: trajectory} shows the trajectories of a participant under varying delay conditions, compared to the baseline path. Where MLPD quantifies the deviation degree between the robot's actual trajectory and the baseline path.

\subsubsection{EEG data}
The raw EEG data were contaminated by artifacts including eye blinks and muscle activities, which required identification and removal. The EEGLAB toolbox was employed for preprocessing the raw data. After electrode localization, a $1$--$30$ Hz bandpass filter was applied to eliminate noise interference. Subsequently, the EEG signals were decomposed using Independent Component Analysis (ICA), with artifacts automatically identified and removed through the ADJUST plugin. Furthermore, abnormal EEG channels were interpolated using adjacent channels, followed by baseline correction of the EEG data. The above procedures were primarily conducted with reference to established methodologies \cite{yang2025recognizing}.

Following preprocessing, time-domain EEG data were transformed to the frequency domain through Fast Fourier Transform (FFT). The power spectral density was segmented into four bands: $\delta$ ($1$--$3$\,Hz), $\theta$ ($4$--$7$\,Hz), $\alpha$ ($8$--$13$\,Hz) and $\beta$ ($14$--$30$\,Hz). For spatial analysis, $32$ electrode channels were grouped into four regions (frontal, parietal, occipital and temporal) following established protocols \cite{yang2018exploring}, as shown in Fig. \ref{fig: egg}. Mean band power was computed for each region-band combination and converted to decibel (dB) units using logarithmic transformation, yielding a $16$-dimensional feature set ($4$ regions × $4$ bands).

\section{Analysis Methods}
\label{Methods}
\subsection{Robot Behavior Analysis Method}
\begin{figure}[t]
    \centering
    \includegraphics[width=1\linewidth]{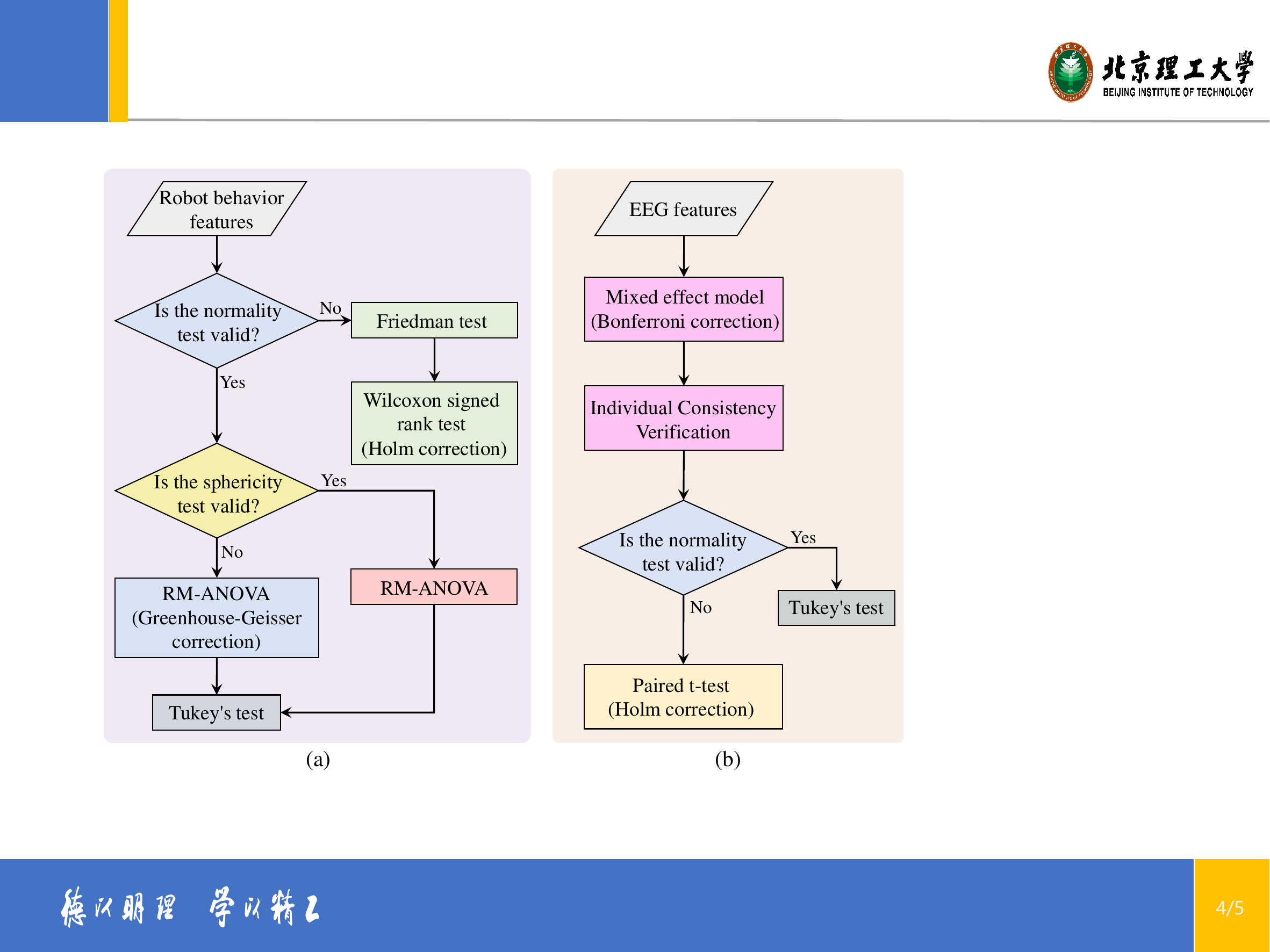}
    \caption{Flowcharts of the (a) robot behavior and (b) EEG analysis procedures.}
    \label{fig: 实验方法new}
\end{figure}

This phase employs repeated-measures analysis of variance (RM-ANOVA) \cite{armstrong2017recommendations} to compare the significance of behavioral differences across groups (with varying delays), with the procedure flow illustrated in Fig. \ref{fig: 实验方法new}(a).

The robot behavior features analysis begins with normality testing (Shapiro-Wilk \cite{razali2011power}). For non-normally distributed data, Friedman's non-parametric test is conducted, followed by post hoc pairwise comparisons with Holm-corrected Wilcoxon signed-rank tests \cite{cleophas2016non}. Normally distributed data proceed to sphericity testing (Mauchly's test \cite{lee2015repeated}), with violations addressed through Greenhouse-Geisser corrected RM-ANOVA and valid sphericity cases using standard RM-ANOVA. Both parametric pathways conclude with Tukey's HSD post-hoc testing \cite{prasad2011financial} for pairwise comparisons.

\subsection{EEG Analysis Method}
Our primary objective at this stage is to identify representative EEG features among $16$ dimensions that effectively characterize teleoperation performance under varying delay conditions. In other words, we seek features demonstrating consistent trends across all participants. The selected features are  subsequently analyzed for between-group significance. Fig. \ref{fig: 实验方法new}(b) shows the procedure flow.

The analysis begins with EEG features, followed by stratified statistical processing. A mixed effect model \cite{sheiner1991introduction} with Bonferroni correction first systematically screens for features exhibiting significant linear/nonlinear trends across groups. The features that pass the screening need to be validated for individual consistency to ensure the generalizability of the response pattern across participants. Subsequently, normality is assessed using Shapiro-Wilk tests. Normally distributed data are analyzed with Tukey's HSD post hoc tests, while non-normal data are evaluated using paired t-tests with Holm correction \cite{niiler2020comparing}.

\section{Results and discussion}
Following the method in Section \ref{Methods}, the corresponding results are presented in Fig. \ref{fig: results} through line graphs and box plots, with detailed descriptions provided below.

\subsection{Robot Behavior Result Analysis}
\subsubsection{Average speed}
\begin{figure*}[t]
    \centering
    \includegraphics[width=0.95\linewidth]{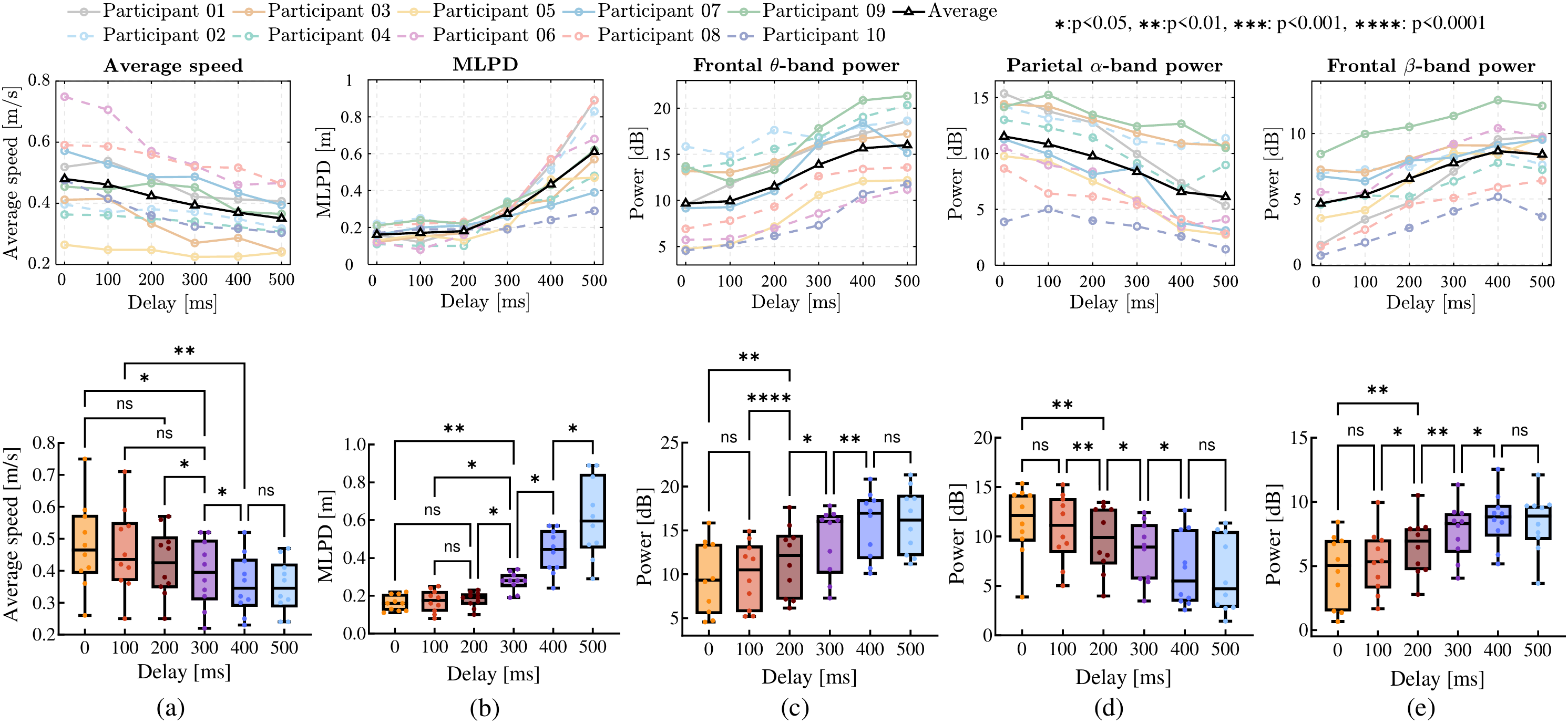}
    \caption{Statistical analysis results of robot behavior and EEG features, including line graphs (the first row) and box plots (the second row).}
    \label{fig: results}
\end{figure*}

The average speed data across different delay conditions pass normality tests ($p$-values: $0.969$, $0.956$, $0.619$, $0.602$, $0.995$, $0.528$; all $>0.05$). Mauchly's test indicates violation of sphericity ($W=0.12$, $p<0.05$), and Greenhouse-Geisser corrected RM-ANOVA ($\varepsilon=0.31$) reveals a significant main effect of delay level on speed ($F(1.52,13.71)=18.67$, $p<0.001$, $\eta^2=0.67$). The line graph in Fig. \ref{fig: results}(a) clearly shows that participants significantly reduce the mobile robot's control speed with increasing delays. This speed reduction strategy reflects \textbf{participants' deliberate compensation for instability caused by time delays, sacrificing speed for enhanced control precision.}

Post-hoc Tukey tests (box plot in Fig. \ref{fig: results}(a)) demonstrate the first statistically significant speed reduction between 200 ms and 300 ms delay conditions ($p=0.01$, $95$\% CI: $[0.008, 0.056]$), establishing \textbf{the delay sensitivity threshold for teleoperation efficiency within the 200--300\,ms range.} 
Notably, the nonsignificant difference between 400-500 ms delays ($p=0.25$, $95$\% CI: $[-0.009, 0.047]$) suggests \textbf{400 ms likely represents the compensation threshold, beyond which speed adjustments cease.}

\subsubsection{MLPD}
The MLPD data satisfy normality assumptions (\emph{p}-values: $0.188$, $0.735$, $0.527$, $0.216$, $0.441$, $0.551$; all $>0.05$). Despite violating the sphericity assumption ($W=0.15$, $p<0.05$), Greenhouse-Geisser corrected RM-ANOVA ($\varepsilon=0.29$) reveals a significant main effect of delay level on MLPD ($F(1.45,13.01)=39.3$, $p<0.001$, $\eta^2=0.81$). The line graph in Fig. \ref{fig: results}(b) demonstrates a progressive increase in MLPD with higher system delays, confirming the \textbf{negative effect of delay on teleoperation accuracy.}

Post-hoc Tukey tests (box plot in Fig. \ref{fig: results}(b)) reveal the first significant MLPD increase between 200 ms and 300 ms ($p = 0.011$, $95$\% CI $[-0.17, -0.02]$), indicating \textbf{this range as the critical delay threshold for accuracy decline in teleoperation.} Beyond this threshold, compensation ability decreases significantly, leading to progressive human performance deterioration.

\subsection{EEG Result Analysis}

The mixed effect model identifies three delay-dependent EEG features (Fig. \ref{fig: results}(c)-(e)): frontal $\theta$- and $\beta$-band power increase with delay duration, while parietal $\alpha$-band power decreases. Prior cognitive neuroscience studies have demonstrated that elevated cognitive workload induces similar spectral changes in these frequency bands. These findings indicate that \textbf{system delays significantly increase participants' cognitive workload.}

\begin{table}[t]
\centering
\caption{\textsc{The Results of normality tests for EEG features}}
\label{tab:normality tests} 
\renewcommand{\arraystretch}{1.2}
\setlength{\tabcolsep}{4pt} % 缩小列间距
\begin{tabular}{lcccccc}
\hline\hline
& \multicolumn{6}{c}{Delay (ms)} \\ 
\cline{2-7}
EEG features & 0  & 100 & 200 & 300 & 400 & 500 \\ 
\hline
Frontal $\theta$-band power & $0.274$ & $0.233$ & $0.489$ & $0.061$ & $0.333$ & $0.375$ \\
Frontal $\beta$-band power & $0.361$ & $0.963$ & $0.799$ & $0.844$ & $0.803$ & $0.612$ \\
Parietal $\alpha$-band power & $0.208$ & $0.556$ & $0.195$ & $0.487$ & $0.109$ & $0.078$ \\
\hline\hline
\end{tabular}
\end{table}

To further analyze between-group differences under varying delays, we first conduct normality tests on three EEG features (Table \ref{tab:normality tests}), confirming their normal distribution (all $p$-values $>0.05$). Based on this, we perform Tukey's post hoc test for multiple comparisons. The results reveal that the three features first exhibit significant differences between 100 ms and 200 ms, while no statistical differences are observed between 400 ms and 500 ms (Table \ref{tab:significant differences}). These findings suggest that \textbf{human perceptual sensitivity to delay exhibits a threshold window of 100-200 ms,} and \textbf{the ceiling of cognitive resource allocation may be 400 ms,} as the neural responses tend to stabilize.

\begin{table}[htp]
\centering
\caption{\textsc{The Results of significant differences for EEG Features}}
\label{tab:significant differences} 
\renewcommand{\arraystretch}{1.3}
\setlength{\tabcolsep}{3pt} % 缩小列间距
\begin{tabular}{lcc}
\hline\hline
& \multicolumn{2}{c}{Delay (ms)} \\ 
\cline{2-3}
EEG features & 100 vs. 200 & 400 vs. 500 \\ 
\hline
Frontal $\theta$-band power & 
\makecell[t]{$p < 0.001$ \\ (95\% CI: [-2.22, -2.04])} & 
\makecell[t]{$p = 0.95$ \\ (95\% CI: [-1.85, 1.16])} \\
Frontal $\beta$-band power & 
\makecell[t]{$p = 0.01$ \\ (95\% CI: [-2.22, -0.26])} & 
\makecell[t]{$p = 0.92$ \\ (95\% CI: [-0.69, 1.21])} \\
Parietal $\alpha$-band power & 
\makecell[t]{$p = 0.002$ \\ (95\% CI: [0.43, 1.72])} & 
\makecell[t]{$p = 0.87$ \\ (95\% CI: [-1.02, 1.94])} \\
\hline\hline
\end{tabular}
\end{table}

\section{Conclusion}
This study reveals the effects of communication delay on human performance and neurocognitive responses in mobile robot teleoperation through a human-in-the-loop simulation experiment, including
\begin{itemize}
    \item Human performance: Increased delay negatively impacts robot efficiency (average speed) and accuracy (MLPD), with statistically significant degradation first observed in the 200-300 ms delay range.
    \item Neurocognitive responses: EEG features, including frontal $\theta$/$\beta$-band and parietal $\alpha$-band power, exhibit significant delay dependence. We further identify the early physiological perception threshold range for delay (100-200 ms) and the cognitive resource allocation limit (400 ms) in teleoperation tasks.
\end{itemize}

These findings can guide the design of safe and reliable teleoperation systems by suggesting parameters such as a 400 ms maximum allowable delay threshold and early-warning mechanisms for the 100–200 ms range. Future work will develop task-adaptive delay prediction-compensation algorithms using multimodal physiological signals, dynamically linking task complexity to neurobehavioral thresholds to optimize cognitive load in human-machine systems.

\section*{Acknowledgment}

This work was supported by Diesel Engine Development Project (DLZX202306).

\bibliographystyle{IEEEtran}
% \bibliography{IEEEtranBST/IEEEexample}
\bibliography{IEEEexample}

\begin{thebibliography}{10}
\providecommand{\url}[1]{#1}
\csname url@samestyle\endcsname
\providecommand{\newblock}{\relax}
\providecommand{\bibinfo}[2]{#2}
\providecommand{\BIBentrySTDinterwordspacing}{\spaceskip=0pt\relax}
\providecommand{\BIBentryALTinterwordstretchfactor}{4}
\providecommand{\BIBentryALTinterwordspacing}{\spaceskip=\fontdimen2\font plus
\BIBentryALTinterwordstretchfactor\fontdimen3\font minus \fontdimen4\font\relax}
\providecommand{\BIBforeignlanguage}[2]{{%
\expandafter\ifx\csname l@#1\endcsname\relax
\typeout{** WARNING: IEEEtran.bst: No hyphenation pattern has been}%
\typeout{** loaded for the language `#1'. Using the pattern for}%
\typeout{** the default language instead.}%
\else
\language=\csname l@#1\endcsname
\fi
#2}}
\providecommand{\BIBdecl}{\relax}
\BIBdecl

\bibitem{lichiardopol2007survey}
S.~Lichiardopol, ``A survey on teleoperation,'' 2007.

\bibitem{aguilar2024development}
I.~Aguilar and D.~Barrios, ``Development of a teleoperation system for controlling a scaled prototype robot for rescue tasks in natural disasters,'' in \emph{2024 Latin American Robotics Symposium (LARS)}.\hskip 1em plus 0.5em minus 0.4em\relax IEEE, 2024, pp. 1--6.

\bibitem{abubakar2024predictor}
A.~Abubakar, Y.~Zweiri, R.~Alhammadi, M.~B. Mohiuddin, M.~Yakubu, and L.~Seneviratne, ``Predictor-based control for delay compensation in bilateral teleoperation of wheeled rovers on soft terrains,'' \emph{IEEE Access}, 2024.

\bibitem{kot2018application}
T.~Kot and P.~Nov{\'a}k, ``Application of virtual reality in teleoperation of the military mobile robotic system taros,'' \emph{International journal of advanced robotic systems}, vol.~15, no.~1, p. 1729881417751545, 2018.

\bibitem{kamtam2024network}
S.~B. Kamtam, Q.~Lu, F.~Bouali, O.~C. Haas, and S.~Birrell, ``Network latency in teleoperation of connected and autonomous vehicles: A review of trends, challenges, and mitigation strategies,'' \emph{Sensors}, vol.~24, no.~12, p. 3957, 2024.

\bibitem{kim2021impact}
E.~Kim, V.~Peysakhovich, and R.~N. Roy, ``Impact of communication delay and temporal sensitivity on perceived workload and teleoperation performance,'' in \emph{ACM symposium on applied perception 2021}, 2021, pp. 1--8.

\bibitem{lu2019workload}
S.~Lu, M.~Y. Zhang, T.~Ersal, and X.~J. Yang, ``Workload management in teleoperation of unmanned ground vehicles: Effects of a delay compensation aid on human operators’ workload and teleoperation performance,'' \emph{International Journal of Human--Computer Interaction}, vol.~35, no.~19, pp. 1820--1830, 2019.

\bibitem{blackett2021effects}
C.~Blackett, A.~Fernandes, E.~Teigen, and T.~Thoresen, ``Effects of signal latency on human performance in teleoperations,'' in \emph{International Conference on Human Interaction and Emerging Technologies}.\hskip 1em plus 0.5em minus 0.4em\relax Springer, 2021, pp. 386--393.

\bibitem{zheng2020evaluation}
Y.~Zheng, M.~J. Brudnak, P.~Jayakumar, J.~L. Stein, and T.~Ersal, ``Evaluation of a predictor-based framework in high-speed teleoperated military ugvs,'' \emph{IEEE Transactions on Human-Machine Systems}, vol.~50, no.~6, pp. 561--572, 2020.

\bibitem{lim2025communication}
T.~Lim, M.~Hwang, J.~Byeon, J.~An, S.~Park, H.~Cha, S.~Yoon, and E.~Kim, ``Communication delay thresholds for effective teleoperation in a mobility system,'' \emph{International Journal of Human--Computer Interaction}, pp. 1--17, 2025.

\bibitem{neumeier2019teleoperation}
S.~Neumeier, P.~Wintersberger, A.-K. Frison, A.~Becher, C.~Facchi, and A.~Riener, ``Teleoperation: The holy grail to solve problems of automated driving? sure, but latency matters,'' in \emph{Proceedings of the 11th International Conference on Automotive User Interfaces and Interactive Vehicular Applications}, 2019, pp. 186--197.

\bibitem{ridderinkhof2014neurocognitive}
K.~R. Ridderinkhof, ``Neurocognitive mechanisms of perception--action coordination: A review and theoretical integration,'' \emph{Neuroscience \& Biobehavioral Reviews}, vol.~46, pp. 3--29, 2014.

\bibitem{lee2022operator}
D.-H. Lee, ``Operator-centric joystick mapping for intuitive manual operation of differential drive robots,'' \emph{Computers and Electrical Engineering}, vol. 104, p. 108427, 2022.

\bibitem{yang2025recognizing}
L.~Yang, R.~Zhou, G.~Li, Y.~Yang, and Q.~Zhao, ``Recognizing and explaining driving stress using a shapley additive explanation model by fusing eeg and behavior signals,'' \emph{Accident Analysis \& Prevention}, vol. 209, p. 107835, 2025.

\bibitem{yang2018exploring}
L.~Yang, Z.~He, W.~Guan, and S.~Jiang, ``Exploring the relationship between electroencephalography (eeg) and ordinary driving behavior: a simulated driving study,'' \emph{Transportation research record}, vol. 2672, no.~37, pp. 172--180, 2018.

\bibitem{armstrong2017recommendations}
R.~A. Armstrong, ``Recommendations for analysis of repeated-measures designs: testing and correcting for sphericity and use of manova and mixed model analysis,'' \emph{Ophthalmic and Physiological Optics}, vol.~37, no.~5, pp. 585--593, 2017.

\bibitem{razali2011power}
N.~M. Razali, Y.~B. Wah \emph{et~al.}, ``Power comparisons of shapiro-wilk, kolmogorov-smirnov, lilliefors and anderson-darling tests,'' \emph{Journal of statistical modeling and analytics}, vol.~2, no.~1, pp. 21--33, 2011.

\bibitem{cleophas2016non}
T.~J. Cleophas, A.~H. Zwinderman, T.~J. Cleophas, and A.~H. Zwinderman, ``Non-parametric tests for three or more samples (friedman and kruskal-wallis),'' \emph{Clinical data analysis on a pocket calculator: understanding the scientific methods of statistical reasoning and hypothesis testing}, pp. 193--197, 2016.

\bibitem{lee2015repeated}
Y.~Lee, ``What repeated measures analysis of variances really tells us,'' \emph{Korean journal of anesthesiology}, vol.~68, no.~4, pp. 340--345, 2015.

\bibitem{prasad2011financial}
K.~Prasad and A.~Chari, ``Financial performance of public and private sector banks: an application of post-hoc tukey hsd test,'' \emph{Indian Journal of Commerce and Management Studies}, vol.~2, no.~5, pp. 79--92, 2011.

\bibitem{sheiner1991introduction}
L.~B. Sheiner and T.~H. Grasela, ``An introduction to mixed effect modeling: concepts, definitions, and justification,'' \emph{Journal of pharmacokinetics and biopharmaceutics}, vol.~19, pp. S11--S24, 1991.

\bibitem{niiler2020comparing}
T.~Niiler, ``Comparing groups of time dependent data using locally weighted scatterplot smoothing alpha-adjusted serial t-tests,'' \emph{Gait \& Posture}, vol.~76, pp. 58--63, 2020.

\end{thebibliography}

\end{document}